\documentclass[runningheads]{llncs}
\usepackage{mathrsfs}
\usepackage{amsfonts}
\usepackage{xcolor}
\usepackage[pdftex]{graphicx}
\usepackage{caption}
\usepackage{amssymb}
\usepackage{mathrsfs}
\usepackage{colortbl}
\usepackage{amsfonts}
\usepackage{adjustbox}
\usepackage{url}
\usepackage{multirow}
\usepackage{booktabs}
\usepackage{amsmath}
\usepackage{caption}
\usepackage{makecell}
\usepackage{subfigure}
\usepackage{mathrsfs}
\usepackage{color}
\usepackage{booktabs}
\usepackage{array}
\usepackage{marvosym}
\usepackage{colortbl}
\definecolor{headerbg}{RGB}{240,240,240}
\usepackage{amssymb}

\usepackage[T1]{fontenc}
\usepackage{graphicx}
\begin{document}
\title{A Frequency-Aware Self-Supervised Learning for Ultra-Wide-Field Image Enhancement}
%
\author{Weicheng Liao\inst{1,2}\textsuperscript{\dag} \and 
Zan Chen\inst{1}\textsuperscript{\dag} \and
Jianyang Xie\inst{3} \and
Yalin Zheng\inst{3} \and
Yuhui Ma\inst{2}\textsuperscript{(\Letter)} \and
Yitian Zhao\inst{2,4}\textsuperscript{(\Letter)}}
\authorrunning{W. Liao et al.}
%
\titlerunning{Frequency-Aware UWF Image Enhancement}
\institute{College of Information Engineering, Zhejiang University of Technology, Hangzhou, China \and
Ningbo Institute of Materials Technology and Engineering, Chinese Academy of Sciences, Ningbo, China\\
\email{\{mayuhui, yitian.zhao\}@nimte.ac.cn}\\
 \and
Department of Eye and Vision Science, University of Liverpool, Liverpool, UK \and
School of Biomedical Engineering, ShanghaiTech University, Shanghai, China\\
}

\maketitle              
\makeatletter
\renewcommand{\@makefntext}[1]{\noindent #1}
\makeatother
\footnotetext{\textsuperscript{\dag}~These authors contributed equally.}
\begin{abstract}
Ultra-Wide-Field (UWF) retinal imaging has revolutionized retinal diagnostics by providing a comprehensive view of the retina. However, it often suffers from quality-degrading factors such as blurring and uneven illumination, which obscure fine details and mask pathological information. While numerous retinal image enhancement methods have been proposed for other fundus imageries, they often fail to address the unique requirements in UWF, particularly the need to preserve pathological details. In this paper, we propose a novel frequency-aware self-supervised learning method for UWF image enhancement. It incorporates frequency-decoupled image deblurring and Retinex-guided illumination compensation modules. An asymmetric channel integration operation is introduced in the former module, so as to combine global and local views by leveraging high- and low-frequency information, ensuring the preservation of fine and broader structural details. In addition, a color preservation unit is proposed in the latter Retinex-based module, to provide multi-scale spatial and frequency information, enabling accurate illumination estimation and correction. Experimental results demonstrate that the proposed work not only enhances visualization quality but also improves disease diagnosis performance by restoring and correcting fine local details and uneven intensity. To the best of our knowledge, this work is the first attempt for UWF image enhancement, offering a robust and clinically valuable tool for improving retinal disease management.

\end{abstract}
\section{Introduction}

\begin{figure}[t]
\centering{
\includegraphics[width=11cm]{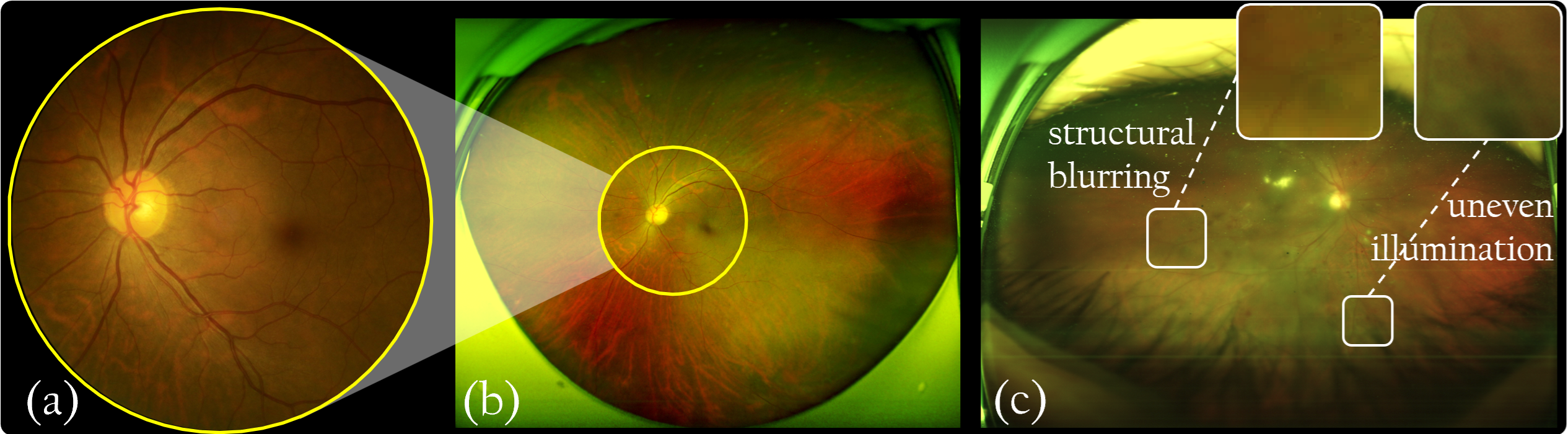}
}
\caption{Illustrations of UWF imaging. (a-b) A color fundus image and an UWF image acquired from the same eye. (c) Sample low-quality factors observed in UWF image: vasculature blurring and uneven illumination mask the lesion.}
\label{fig.0}
\end{figure}

Ultra-Wide-Field (UWF) retinal imaging represents a significant advancement in retinal diagnostics, offering a broader and more detailed view of the retina compared with conventional imaging approaches. Fig.~\ref{fig.0} (a-b) illustrates this by comparing a retinal color fundus image with a UWF image,  highlighting the enhanced ability of UWF imaging to capture a more comprehensive retina, improving the diagnosis and management of various retinal diseases.

In recent years, numerous image enhancement methods have been proposed~\cite{dong2023multi,wang2024zero,wu2023learning,xu2023low} to improve image readability. However, the majority of these methods are primarily designed for natural images and fail to adequately address the specific requirements of medical imaging, particularly retinal imaging, where preserving micro pathological details is critical. Numerous retinal image enhancement methods have been introduced that learn general priors from extensive datasets comprising low- and high-quality images~\cite{9810184,lee2023deep}. However, acquiring such paired datasets poses significant challenges in medical imaging. Consequently, researchers~\cite{perez2020conditional,ma2020cycle,9288835,liu2022degradation} have synthesized low-quality images by degrading high-quality ones, thereby constructing paired datasets for training. 
Nevertheless, these approaches rely heavily on the availability of extensive collections of high-quality images free from quality-degrading factors. 

The UWF enhancement is more complicated by several low-quality factors, with blurring and uneven illumination being the most significant. Blurring, as shown in Fig.~\ref{fig.0} (c), often obscures fine local details, reducing the diagnostic utility of the images. Similarly, uneven illumination, depicted in Fig.~\ref{fig.0} (c), particularly in the peripheral retinal regions, the poor contrast may mask critical pathological information. The aforementioned issues collectively hinder the identification of important structural features, making it more difficult for clinicians to accurately and confidently interpret the images. UWF images are acquired through red and green laser scanning, which means they do not follow the "gray world assumption" and are thus prone to color distortion during light compensation~\cite{4635754}. In addition to blur and uneven illumination, the preservation of pathological details, peripheral distortions obtained by unsupervised approach, and the lack of paired datasets for training. 
Overcoming these challenges is essential to ensure that enhanced UWF images provide reliable and clinically useful information for diagnosing and managing retinal diseases.

In this work, we introduce a frequency-aware self-supervised learning method, which is specifically designed for UWF image enhancement. The method consists of a FREquency-Decoupled deblurring (FRED) module and a Retinex-guided Illumination CompEnsation (RICE) module. In the FRED module, an asymmetric channel integration operation is introduced, to effectively combine global and local views by leveraging separated high- and low-frequency information. This ensures the preservation of both fine and broader structural features. In the RICE module, a color preservation unit is proposed to provide multi-scale spatial and frequency information, enabling accurate illumination estimation and correction. Together, these innovations allow the model to restore fine local details and correct intensity inhomogeneities while preserving structural integrity. 

The main contributions of this paper are summarized as follows:~\textbf{1)} To the best of our knowledge, this work represents the first dedicated effort for deblurring and illumination compensation in UWF image enhancement. The experimental findings demonstrate that the work not only enhances the visual quality but also improves the performance of the subsequent disease diagnosis.~\textbf{2)} The asymmetric channel integration unit effectively combines global and local views by leveraging separated high- and low-frequency information, to ensure the preservation of both fine details and broader structural features.~\textbf{3)} The color preservation unit is able to provide multi-scale spatial and frequency information, enabling accurate illumination estimation and correction.

\section{Proposed Method}

\begin{figure}[t]
\centering{
\includegraphics[width=12cm]{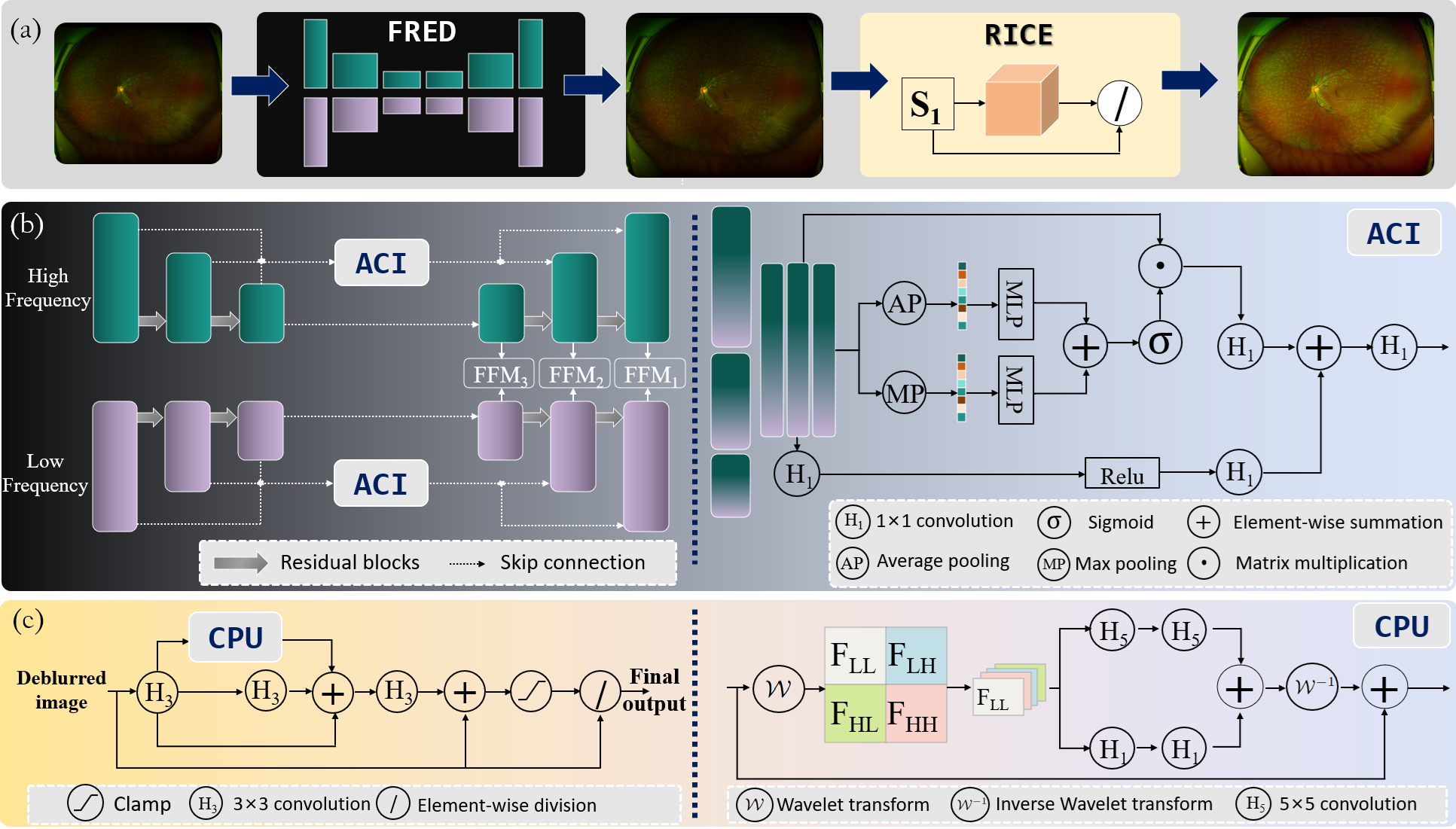}
}
\caption{Overview of the proposed framework and its main components. (a) Overall architecture: the low-quality images are sequentially passed through the deblurring and illumination compensation modules. (b-c) Architectures of FRED and RICE. }
\label{fig.1}
\end{figure}

Fig.~\ref{fig.1} (a) illustrates the overview of the proposed framework, where the image is processed in two stages. In the first stage, the FRED module is proposed to perform deblurring on UWF images, which aims to align the image domain with clinical acquisition conditions of UWF fundus cameras for clearer details. In the second stage, the RICE module is proposed for further illumination compensation based on the deblurring results, which aims at better quality of UWF images in terms of clarity and uniform intensity. 

\subsection{Frequency-decoupled image deblurring (FRED)}


In the deblurring stage, a random blurring degradation operation~\cite{perez2020conditional} is firstly applied to high-quality images with clear structures to generate paired blur/clear samples.  These paired samples are then used to train the model, enabling it to learn an effective restoration mapping and effectively deblur UWF images while preserving fine structural details. In order to prevent mutual interference between varying frequencies, we use average pooling separation (APS)~\cite{yang2024crnet} to decompose the UWF image into high- and low-frequency features, which are processed separately through dual frequency streams with encoder-decoder structure. 
Specifically, the generated blurry image is firstly fed into the APS to obtain the low-frequency features via downsampling by an average pooling layer and upsampling by bilinear interpolation successively. Then the high-frequency features are obtained by substraction of the low-frequency features from the input blurry image. 
Next, both high- and low-frequency features are downsampled to multiple scales and then fed into different encoder layers of dual frequency streams respectively, which have the same encoder-decoder structure.  
In order to effectively aggregate global and local information in both high- and low-frequency features, an Asymmetric Channel Integration (ACI) unit is proposed for cross-scale latent feature fusion in the skip connections of each frequency stream, as shown in Fig.~\ref{fig.1} (b). 





\noindent \textbf{Asymmetric Channel Integration (ACI)}
The proposed ACI unit aims to achieve more comprehensive fusion of global and local features from different encoder layers and further enhance feature representations in the decoder. 
Specifically, the features from different encoder layers are firstly rescaled to corresponding decoder feature size and concatenated to obtain the fusion feature $f_{s} \in \mathbb{R}^{H \times W \times C}$. Then $f_{s}$ is input into global information extraction branch and local information extraction branch, respectively. Global information extraction branch adopts channel-wise attention learning to refine $f_{s}$. In this branch, max and average pooling operations followed by a Multi Layer Perceptron (MLP) are separately applied to $f_{s}$ to produce two feature vectors $f_{s}^{m}, f_{s}^{a} \in \mathbb{R}^{C}$. Then the two feature vectors are fused via summation and Sigmoid to generate the final attention weight $W_s^g \in \mathbb{R}^{C}$, which is multiplied with $f_{s}$ in an channel-wise manner followed by a $1 \times 1$ convolutional layer to obtain the global information $f_{g}$. For local information extraction branch, two $1 \times 1$ convolutional layers with ReLU is applied to $f_{s}$ to obtain the local information $f_{l}$. Finally, a $1 \times 1$ convolutional layer is applied to summation of $f_{g}$ and $f_{l}$ to calculate the output feature $f_{ACI}$ of the ACI unit, which will be integrated into the corresponding decoder layer. 
The whole process can be formulated as: 

\begin{equation}\small
\begin{gathered}
f_{s}^{a} = M(AP(f_s)), f_{s}^{m} = M(MP(f_s)),\\
W_s^g = \sigma(f_{s}^{a} + f_{s}^{m}), f_{g} = H_1(W_s^g \times f_s),\\
f_{l} = H_1(\delta(H1(f_s))), 
f_{ACI} = H_{1}(f_{g} + f_{l}),
\end{gathered}
\end{equation}
where $M$, $AP$, $MP$, $\delta$, $\sigma$ and $H_{1}$ represent MLP, average pooling, max pooling, ReLU, Sigmoid and $1 \times 1$ convolution, respectively. 

The FRED module further integrates features of each decoder layer separately in high- and low-frequency streams through Frequency Fusion module (FFM)~\cite{yang2024crnet} to output multi-scale deblurring results, as illustrated in Fig.~\ref{fig.1} (b). Each FFM is composed of convolutional blocks and channel attention mechanisms, which facilitates the integration of high- and low-frequency information. The output of FFM on the last decoder layer is the final deblurring result. The FRED module adopted the deep supervision strategy for training and is optimized through multi-scale content loss $L_{cont}$, multi-scale frequency reconstruction loss $L_{MSFR}$~\cite{Cho_2021_ICCV} and perceptual loss $L_{per}$~\cite{zhang2022deep}, which can be formulated as: 
\begin{equation}\small
\begin{aligned}
&L_{deblur} = L_{cont} + \beta L_{MSFR} + \gamma L_{per}
\end{aligned}
\end{equation}
where $\beta$ and $\gamma$ are weighting factors of multi-scale frequency reconstruction loss and perceptual loss, respectively. 



\subsection{Retinex-based illumination compensation (RICE)}

Inspired by SCI~\cite{Ma_2022_CVPR}, we propose a Retinex-based module called RICE for further illumination compensation of the deblurred UWF images, as shown in Fig.~\ref{fig.1} (c). According to the Retinex theory~\cite{Ma_2022_CVPR}, a given image $\textbf{I}$ can be decomposed into two components, respectively the reflectance $\textbf{R}$ and illumination $\textbf{L}$: $\textbf{I} = \textbf{R} \otimes \textbf{L}$. Similarly, the enhanced image $\textbf{I'}$ can also be expressed as $\textbf{I'} = \textbf{R'} \otimes \textbf{L'}$. For the enhancement task that requires only illumination compensation, we can assume that the reflectance should remain constant after enhancement, i.e., $\textbf{R'} = \textbf{R}$. Therefore, the enhanced image $\textbf{I'}$ can be further formulated as $\textbf{I'} = \textbf{R'} \otimes \textbf{L} \otimes (\textbf{L'} \oslash \textbf{L}) = \textbf{R} \otimes \textbf{L} \oslash (\textbf{L} \oslash \textbf{L'}) = \textbf{I} \oslash \textbf{r}$, where $\textbf{r} = \textbf{L} \oslash \textbf{L'}$ represents the ratio of illumination compensation. The proposed RICE aims to precisely estimate the ratio $\textbf{r}$ to achieve the effective illumination compensation.  

However, due to the optical nature of ultra-wide angle fundus imaging, which is likely to raise color distortion for Retinex-based methods during illumination compensation~\cite{4635754}. To address this issue, we propose a Color Preservation Unit (CPU) that incorporates a two-dimensional discrete wavelet transform to ensure color fidelity, as shown in Fig.~\ref{fig.1} (c). This unit utilizes the multi-scale frequency separation capability of the wavelet transform to decouple chromaticity and luminance, leveraging the low-frequency sub-bands to focus on light estimation and suppressing the interference of high-frequency chromaticity information.


\noindent\textbf{Color Preservation Unit (CPU)}
The discrete wavelet transform $\mathcal{W}$ involved in the proposed CPU utilizes four filters defined as: 
\begin{equation}\small
\begin{gathered}
f^{LL}=\begin{pmatrix}1&1\\1&1\end{pmatrix}, 
f^{LH}=\begin{pmatrix}-1&-1\\1&1\end{pmatrix}, f^{HL}=\begin{pmatrix}-1&1\\-1&1\end{pmatrix}, f^{HH}=\begin{pmatrix}1&-1\\-1&1\end{pmatrix}. 
\end{gathered}
\end{equation}
Through the discrete wavelet transform $\mathcal{W}$, the feature $F\in \mathbb{R}^{H \times W \times C}$ of the deblurred UWF image extracted by a $3 \times 3$ convolutional layer is decomposed into four frequency sub-bands: $F^{LL}, F^{LH}, F^{HL}, F^{HH} \in \mathbb{R}^{\frac{H}{2} \times \frac{W}{2} \times C}$, where $F^{LL}$ mainly contains the global features and approximate contours of the image, $F^{LH}$, $F^{HL}$ and $F^{HH}$ capture the details and edge variations of the image. 
Each frequency sub-band is input to two parallel branches respectively composed of two successive $1 \times 1$ and $5 \times 5$ convolutional layers, then the output feature maps of the two branches are fused via an element-wise summation. Finally, the discrete wavelet inverse transform $\mathcal{W}^{-1}$ is applied to these processed frequency sub-bands, then a shortcut connection of the feature $F$ is introduced to produce the output of CPU. 
Let set $\Tilde{F} =\{F^{LL}, F^{LH}, F^{HL}, F^{HH}\}$, and the whole process can be defined as: 
\begin{equation}\small
\begin{gathered}
\Tilde{F} = \mathcal{W}(F),\\
F_{w} = \{H_{5}^2(F^*) + H_{1}^2(F^*) \mid  F^* \in \Tilde{F} \}, \\
F_{wt} = \mathcal{W}^{-1}(F_{w}) + F,
\end{gathered}
\end{equation}
where $H_{1}^2$ and $H_{5}^2$ denote two successive $1 \times 1$ and $5 \times 5$ convolutional layers, respectively, $F_{w}$ and $F_{wt}$ denote the set of the processed frequency sub-bands and the output of CPU, respectively. 
For effective optimization of the RICE module, we adopt a joint loss function containing fidelity loss $L_{f}$, smoothness loss $L_{s}$~\cite{Ma_2022_CVPR} and exposure control loss $L_{exp}$~\cite{Guo_2020_CVPR}, which can be formulated as: 
\begin{equation}\small
\begin{aligned}
&L_{ic} = \alpha L_{f} + L_{s} + L_{exp},
\end{aligned}
\end{equation}
where $\alpha$ is the parameter to control the weight of fidelity loss.

\section{Experimental Results}

\textbf{Dataset description} In this study, we first construct a UWF image enhancement dataset, which involves both healthy controls and patients with diabetic retinopathy (DR). The dataset contains 834 UWF images captured with the Optos 200Tx device, with a resolution of $3900 \times 3072$ pixels and a $200^{\circ}$ field of view. 
In our experiments, 400 images with varying exposure levels were selected for training, while 434 images with different levels of blur and illumination were used to evaluate the performance of the image enhancement method. 
All image acquisitions adhered to the relevant regulatory approvals and patient consent, ensuring compliance with ethical standards and privacy protection protocols.

\noindent \textbf{Experimental settings}
Our method was implemented in Python with PyTorch library. The experiments were carried out on two NVIDIA GPUs (GeForce RTX 4090, 24 GB). All training images were randomly cropped to a size of $256 \times 256$. Adam optimization was adopted with the initial learning rate of 0.0001 and 0.0003 for training the FRED and RICE modules, respectively. The weighted parameters in the loss function were set as: $\beta=0.1$, $\gamma=0.01$, $\alpha=1.5$.

\subsection{Image enhancement performances}

\begin{figure}[t]
\centering{
\includegraphics[width=11cm]{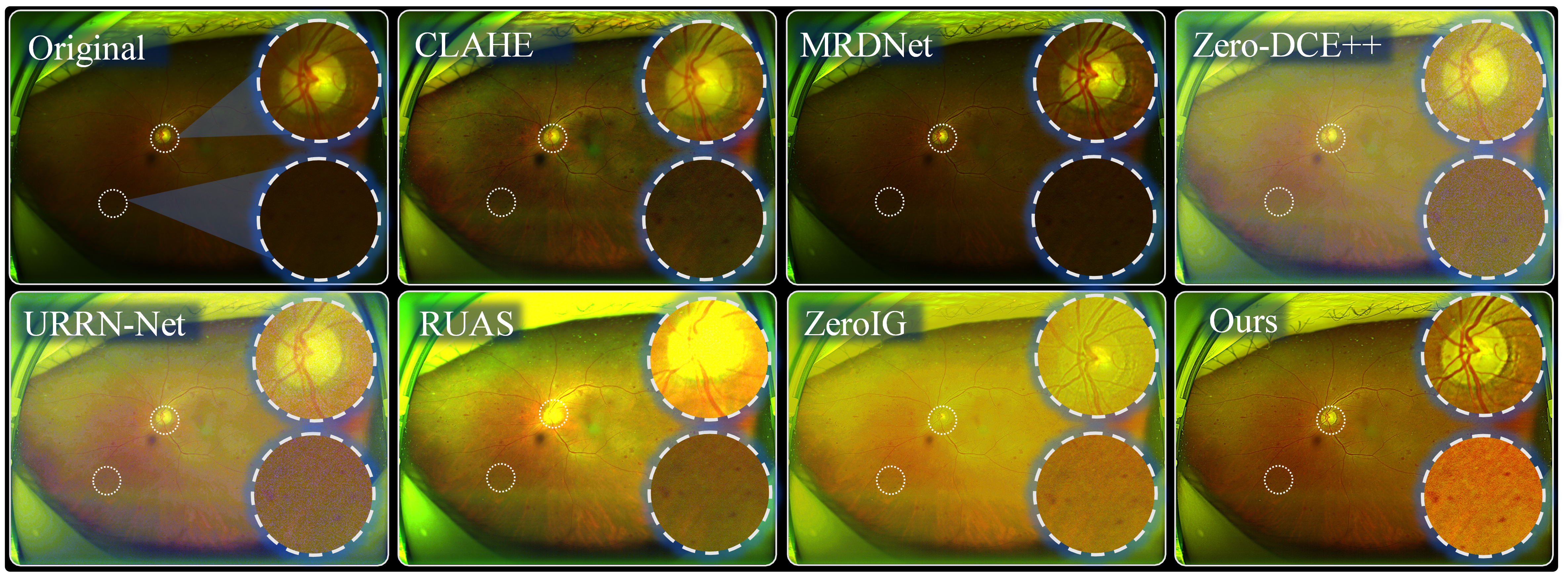}
}
\caption{The visualization results by using different methods.}
\label{fig.2}
\end{figure}

We first present the performance evaluation of the proposed method. Six state-of-the-art methods are used for comparisons, and the NIQE~\cite{mittal2012making}, BRISQUE~\cite{mittal2012no} and PIQE~\cite{venkatanath2015blind} are used as evaluation metrics. Fig.~\ref{fig.2} demonstrates the enhancement results of different methods. It can be observed that several compared methods produce undesirable effects. For example, CLAHE~\cite{zuiderveld1994contrast} amplifies background noise and thus lead to image quality degradation, while both Zero-DCE++~\cite{9369102} and URRN-Net~\cite{CHEN2024105404} generate noticeable light purple artifacts in the peripheral regions, as shown in the bottom-right patch. Deblurring methods such as MRDNet~\cite{zhang2024image} can effectively enhance structural information, but hardly deal with illumination non-uniformity. RUAS~\cite{Liu_2021_CVPR} improves overall brightness while maintaining color consistency, but suffers from overexposure in the optic nerve region, leading to an unnatural appearance. Although ZeroIG~\cite{shi2024zero} effectively enhances overall brightness, it might cause some blurring in fine details, such as capillaries and leopard-like patterns, which can affect the clarity of these regions. In contrast, our method outperforms the others by preserving the continuity of microvessels and maintaining natural color tones. Table~\ref{tab:results} further demonstrate the superiority of our proposed method in quantitative analysis. Our method yields the lowest NIQE, BRISQUE and PIQE scores with large margin when compared with the competing methods. 

\subsection{Ablation study}


In order to investigate the contributions of components in our method, we conducted ablation studies, as illustrated in the bottom four rows of Table~\ref{tab:results}.

\noindent\textbf{Effectiveness of two stages (FRED and RICE)} The results demonstrate a 21.3\% increase in PIQE after removing the FRED module, indicating that intrinsic blur in UWF images has a detrimental effect on illumination compensation, thereby degrading overall image quality. This emphasises the critical role of FRED in eliminating blur, which are essential for image quality enhancement. Furthermore, the absence of the RICE module resulted in a 28\% increase in BRISQUE, confirming the vital role of RICE in ensuring accurate illumination compensation. The absence of RICE led to a failure to balance the overall lighting effect, also leading to a significant degradation in image quality.

\noindent\textbf{Effectiveness of two units (ACI and CPU)} We further validate the effectiveness of ACI in the FRED stage. Disabling ACI resulted in an 8.2\% increase in BRISQUE, indicating the effectiveness of its multi-scale feature fusion within different frequency bands on UWF image deblurring. This might be because the framework without ACI only relies on skip connections for feature fusion, which is less effective in capturing global and local information. 
Similarly, disabling the CPU in the RICE stage led to a 5.4\% increase in NIQE, confirming its significance in maintaining color consistency. The absence of CPU can not effectively extract frequency sub-bands for accurate illumination estimation, thus underperforming in preserving color fidelity during illumination compensation. 

\begin{table}[t]\small
\caption{Image enhancement performance in terms of \textit{quality assessment} (the first three metrics) and \textit{disease grading} tasks (the latter two metrics). }
\label{tab:results}
\centering
   \resizebox{\textwidth}{!}{
    \setlength{\tabcolsep}{3mm}{
\begin{tabular}{lccc||cc}
\toprule
\rowcolor{headerbg}
\textbf{Method} & \textbf{NIQE $\downarrow$} & \textbf{BRISQUE $\downarrow$} & \textbf{PIQE $\downarrow$} & \textbf{ACC(\%) $\uparrow$} & \textbf{F1(\%) $\uparrow$} \\
\midrule
Original & 5.87 $\pm$0.91 & 35.25 $\pm$8.87 & 16.32 $\pm$7.49 & 57.13 & 48.24 \\
CLAHE~\cite{zuiderveld1994contrast} & 4.11 $\pm$0.90 & 22.69 $\pm$8.59 & 10.41 $\pm$5.13 & 61.21 & 53.06 \\
MRDNet~\cite{zhang2024image} & 4.86 $\pm$0.61 & 29.45 $\pm$10.80 & 15.85 $\pm$7.18 & 62.74 & 53.16 \\
Zero-DCE++~\cite{9369102} & 6.80 $\pm$0.53 & 41.12 $\pm$15.12 & 11.75 $\pm$7.52 & 47.81 & 25.86 \\
URRN-Net \cite{CHEN2024105404} & 7.03 $\pm$0.89 & 43.19 $\pm$16.68 & 18.84 $\pm$5.88 & 46.62 & 26.91 \\
RUAS~\cite{Liu_2021_CVPR} & 5.82 $\pm$1.89 & 24.51 $\pm$15.4 & 9.21 $\pm$6.43 & 39.95 & 33.62 \\
ZeroIG~\cite{shi2024zero} & 4.99 $\pm$1.39 & 24.17 $\pm$10.48 & 8.67 $\pm$6.82  & 47.03 & 33.89 \\
\rowcolor{yellow!60}  Ours  & \textbf{4.09 $\pm$0.51} & \textbf{19.15 $\pm$6.02} & \textbf{8.26 $\pm$4.22} & \textbf{69.49} & \textbf{64.07} \\
\hline
\hline
\rowcolor{yellow!40}  w/o FRED & 4.39 $\pm$1.06 & 23.09 $\pm$8.02 & 10.02 $\pm$5.60 & 66.86 & 59.47 \\ 
\rowcolor{yellow!35}  w/o RICE & 4.86 $\pm$0.55 & 24.51 $\pm$9.65 & 14.59 $\pm$7.06 & 62.36 & 51.96 \\
\rowcolor{yellow!30}  w/o ACI & 4.19 $\pm$0.91 & 20.72 $\pm$9.10 & 8.39 $\pm$4.15 & 67.39 & 61.48 \\ 
\rowcolor{yellow!25}  w/o CPU & 4.31 $\pm$0.56 & 20.67 $\pm$7.46 & 8.49 $\pm$4.51 & 68.71 & 62.93 \\
\bottomrule
\end{tabular}}}
\begin{tabular}{p{0.9\textwidth}}
\end{tabular} 
\end{table}

\subsection{Effectiveness of enhancement on disease diagnosis}

To further validate the effectiveness of image enhancement on disease diagnosis, we first collected an additional 902 UWF images (444 normal, 195 mild, 103 moderate, 79 severe non-proliferative DR and 81 proliferative DR), which were split into training and test sets at a 7:3 ratio. Then we trained a ResNet34 model for DR grading using the training set, and employed the well-trained model in the test images enhanced by different methods for comparisons. 
The right two columns of Table~\ref{tab:results} report the DR grading performance on images enhanced by different methods. Our method improves ACC and F1-score by 12.3\% and 15.8\% respectively when compared with the original images, and also outperforms other enhancement methods. This significant improvement is primarily attributed to the FRED and RICE modules, which enhance structural clarity and illumination uniformity, respectively. The findings demonstrate the advantages of our method on improving the visibility of peripheral retinal pathology, thereby facilitating disease diagnosis. 


\section{Conclusions}

In this study, we propose a novel frequency-aware self-supervised learning framework for UWF image enhancement. The primary advantage of our framework is that it can learns to achieve image deblurring and illumination compensation simultaneously via two specially designed modules. First, the FRED module leverages asymmetric channel integration to preserve structural details by combining global and local information in high- and low-frequency information. Second, the RICE module employs a color preservation unit based on frequency decoupling by discrete wavelet transform to mitigate color distortions while achieving illumination uniformity. 
Experimental results demonstrate that our method outperforms other approaches, indicating the potential of the framework in improving image quality for higher accuracy of clinical diagnosis.

\begin{credits}
\subsubsection{\ackname} This study was funded by the National Natural Science Foundation of China (62402478, 62422122, 62272444) and China Postdoctoral Science Foundation (2023M743629)

\subsubsection{\discintname}
The authors have no competing interests to declare that are relevant to the content of this article.
\end{credits}
%
%
%
\bibliographystyle{splncs04}
\bibliography{Paper-2387}

\end{document}